\documentclass{article}

\PassOptionsToPackage{numbers, compress}{natbib}

\usepackage[final]{arxiv}

\usepackage[utf8]{inputenc} 
\usepackage[T1]{fontenc}    
\usepackage{hyperref}       
\usepackage{url}            
\usepackage{booktabs}       
\usepackage{amsfonts}       
\usepackage{microtype}      

\usepackage[cmex10]{amsmath}
\usepackage[table,xcdraw]{xcolor} 
\usepackage{mwe}
\usepackage[caption=false]{subfig}
\usepackage{changepage}
\usepackage{hyperref}

\title{Spatially Supervised Recurrent Convolutional Neural Networks for Visual Object Tracking}

\author{
  Guanghan Ning\thanks{Project Page: 
  	\href{http://guanghan.info/projects/ROLO/}{http://guanghan.info/projects/ROLO/}}, Zhi Zhang, Chen Huang, Zhihai He \\ 
  Department of Electrical and Computer Engineering\\
  University of Missouri\\
  Columbia, MO 65201 \\
  \texttt{ \{gnxr9, zzbhf, chenhuang, hezhi\}@mail.missouri.edu} \\
\And
Xiaobo Ren, Haohong Wang \\
  TCL Research America\\
  \texttt{ \{renxiaobo, haohong.wang\}@tcl.com} \\
}

\begin{document}
\bibliographystyle{plain} 
\maketitle

\begin{abstract}
In this paper, we develop a new approach of spatially supervised recurrent convolutional neural networks for visual object tracking.  Our recurrent convolutional network exploits the history of locations as well as the distinctive visual features learned by the deep neural networks. 
Inspired by recent bounding box regression methods for object detection, we study  the regression capability of Long Short-Term Memory (LSTM) in the temporal domain, and propose to concatenate high-level visual features produced by convolutional networks with region information. In contrast to existing deep learning based trackers that use binary classification for region candidates, we use regression for  direct prediction of the tracking locations both at the convolutional layer and at the recurrent unit.  Our extensive experimental results and performance comparison  with  state-of-the-art tracking methods on challenging benchmark video tracking datasets shows that our tracker is more accurate and robust while maintaining low computational cost. For most test video sequences, our method achieves the best tracking performance, often outperforms the second best by a large margin. 

\end{abstract}

\section{Introduction}
\label{intro}

Visual tracking is a challenging task in computer vision due to target deformations, illumination variations, scale changes,  fast and abrupt motion, partial occlusions, motion blur, object deformation, and background clutters.
Recent advances in methods for object detection \citep{girshick2015fast, ren2015faster} have led to the development of a number of tracking-by-detection \citep{wang2016stct, hall2014categories, hong2015tracking} approaches. These modern trackers are usually complicated systems made up of several separate components. According to \citep{wang2015understanding}, the feature extractor is the most important component of a tracker. Using proper features can dramatically improve the tracking performance. 
To handle tracking failures caused by the above mentioned factors, existing appearance-based tracking methods \citep{dinh2011context, kalal2010pn, henriques2012exploiting} adopt either generative or discriminative models to separate the foreground from background and distinct co-occurring objects. One major drawback is that they rely on low-level handcrafted features which are incapable to capture semantic information of targets, not robust to significant appearance changes, and only have limited discriminative power.
Therefore, more and more trackers are using image features learned by deep convolutional  neural networks \citep{Wang_2015_ICCV, hong2015tracking, wang2013learning}. 
We recognize that existing methods mainly focus on  improving the performance and robustness of deep features against hand-crafted features. How to extend the deep neural network analysis into the spatiotemporal domain for visual object tracking has not been adequately studied. 

In this work, we propose to develop a new  visual tracking approach based on recurrent convolutional neural networks,  which extends the neural network learning and analysis into the spatial and temporal domain. The key motivation behind our method is  that tracking failures can often be effectively recovered by learning from historical visual semantics and tracking proposals. 
In contrast to existing tracking methods based on Kalman filters or related temporal prediction methods, which only consider
the location history, our recurrent convolutional model is “doubly deep” in that it examine the history of locations as well as the robust visual features of past frames.  

There are two  recent papers \citep{kahou2015ratm, gan2015first} that are closely related to this work.
They address  the similar issues of object tracking using recurrent neural networks (RNN), but they focused on artificially generated sequences and synthesized data. The specific challenges of object tracking  in real-world videos have not been carefully addressed. 
They use traditional RNN as an attention scheme to spatially glimpse on different regions and rely on an additional binary classification at local regions. In contrast, we directly regress coordinates or heatmaps instead of using sub-region classifiers. We use the LSTM for an end-to-end spatio-temporal regression with a single evaluation, which proves to be more efficient and effective. 
Our extensive experimental results and performance comparison with state-of-the-art tracking method on challenging benchmark tracking datasets shows that our tracker is more accurate and robust while maintaining low computational cost. For most test sequences, our method achieves the best tracking performance, often outperforms the second best by a large margin. 
     
Major contributions of this work include: 
(1) we introduce a modular neural network that can be trained end-to-end with gradient-based learning methods. Using object tracking as an example application, we explore different settings and provide insights into model design and training, as well as LSTM's interpretation and regression capabilities of high-level visual features.
(2) In contrast to existing ConvNet-based trackers, our proposed framework extends the neural network analysis into the spatiotemporal domain for efficient visual object tracking.
(3) The proposed network is both accurate and efficient with low complexity.

\section{System Overview} \label{system-overview}

\begin{figure}[h] 
      	\centering
      	\captionsetup{justification=centering}
      	\includegraphics[height=2in]{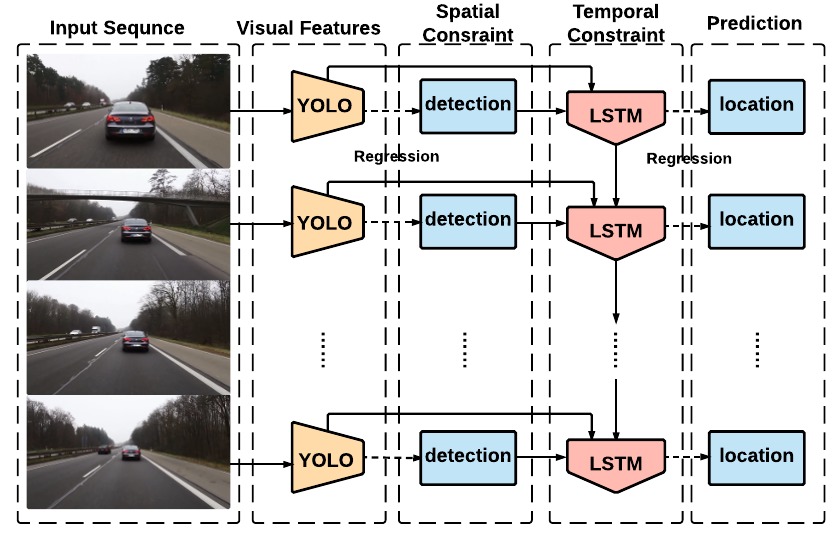}
      	\caption{Simplified overview of our system and the tracking procedure.}
      	\label{fig-overview}
\end{figure}

The overview of the tracking procedures is illustrated in Fig. \ref{fig-overview}. 
We choose YOLO to collect rich and robust visual features, as well as preliminary location inferences; and we use LSTM in the next stage as it is spatially deep and appropriate for sequence processing. 
The proposed model is a deep neural network that takes as input raw video frames and returns the coordinates of a bounding box of an object being tracked in each frame. Mathematically, the proposed model factorizes the full tracking probability into
\begin{equation}
      	p(B_{1}, B_{2}, …, B_{T} | X_{1}, X_{2}, …, X_{T}) = \prod_{t=1}^{T} p(B_{t} | B_{<t}, X_{\leq t}),
\end{equation}
where $B_{t}$ and $X_{t}$ are the location of an object and an input frame, respectively, at time $t$. $B_{<t}$ is the history of all previous locations before time $t$, and $X_{\leq t}$ is the history of input frames up to time $t$. 
In the following section, we describe the major components of the proposed system in more detail.

\subsection{Long Short Term Memory (LSTM)}
Conventional RNNs cannot access long-range context due to the back-propagated error either inflating or decaying over time, which is called the vanishing gradient problem \citep{ hochreiter2001gradient}. 
By contrast, LSTM RNNs \citep{ hochreiter1997long} 
overcome this problem and are able to model self-learned context information. 
The major innovation of LSTM is its memory cell $c_{t}$ which essentially acts as an accumulator of the state information. The cell is accessed, written and cleared by several self-parameterized controlling gates. Every time a new input comes, its information will be accumulated to the cell if the input gate $i_{t}$ is activated. Also, the past cell status $c_{t-1}$ could be “forgotten” in this process if the forget gate $f_{t}$ is on. Whether the latest cell output $c_{t}$ will be propagated to the final state $h_{t}$ is further controlled by the output gate $o_{t}$.
In our system, we use the LSTM unit as the tracking module. 
Unlike standard RNNs, the LSTM architecture uses memory cells to store and output information, allowing it to better discover long-range temporal relations. 
Letting $\sigma = (1+ e^{-x})^{-1}$,  be the sigmoid nonlinearity which squashes real-valued inputs to a $[0, 1]$ range, and letting $\phi(x) = \frac{e^{x} - e^{-x}}{e^{x} + e^{-x}}$, the LSTM updates for timestamp t given inputs $x_{t}$,  $h_{t-1}$, and $c_{t-1}$ are:
\begin{equation}
	\begin{aligned}
		i_{t} &= \sigma(W_{xi}x_{t} + W_{hi}h_{t-1} + b_{i}), \\
		f_{t} &= \sigma(W_{xf}x_{t} + W_{hf}h_{t-1} + b_{f}), \\
		o_{t} &= \sigma(W_{xo}x_{t} + W_{ho}h_{t-1} + b_{o}), \\
		g_{t} &= \sigma(W_{xc}x_{t} + W_{hc}h_{t-1} + b_{c}), \\
		h_{t} &= o_{t} \odot \phi(c_{t}).
	\end{aligned}
\end{equation}

\subsection{Object Detection Using YOLO}

While accuracy is important in visual tracking systems, speed is another significant factor to consider in practice. Existing tracking approaches employing ConvNets are already computationally expensive. Applying it to each frame for visual object tracking will 
result in prohibitively high computational complexity. 
Recently, a new approach to object detection is proposed in \citep{redmon2015you}. They frame object detection as a regression problem to spatially separated bounding boxes and associated class probabilities. The baseline YOLO model processes images in real-time at 45 fps. A smaller version of the network, Fast YOLO, processes at 155 fps while still the state-of-the-art object detection performance.
In one frame, YOLO may output multiple detections. In assigning the correct detection to the tracking target, we employ an assignment cost matrix that is computed as the intersection-over-union (IOU) distance between the current detection and the mean of its short-term history of validated detections. The detection of the first frame, however, is determined by the IOU distance between the detections and the ground truth. Additionally, a minimum IOU is imposed to reject assignments where the detection to target overlap is less than $IOU_{min}$.

\section{Our Proposed System} \label{proposed-system}

\begin{figure}[t] 
	\centering
	\captionsetup{justification=centering}
	\includegraphics[width=5in]{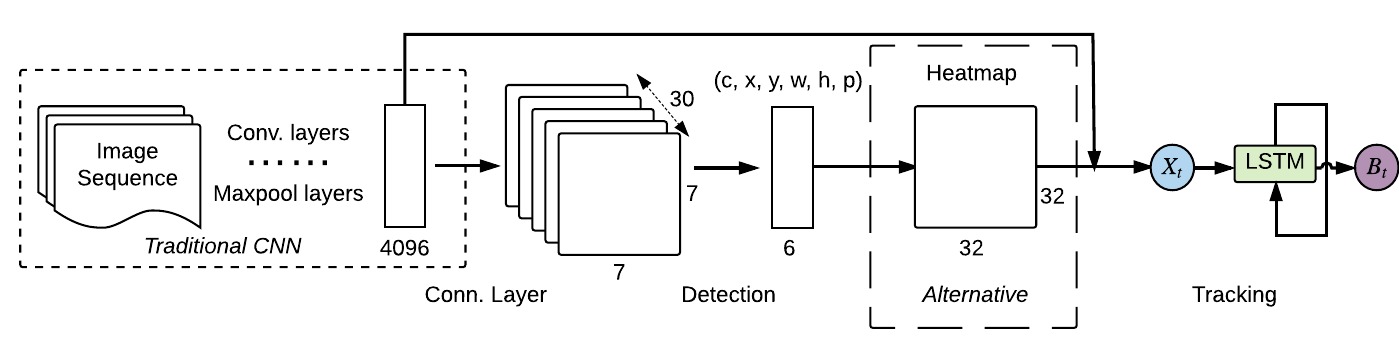}
	\caption{Our proposed architecture.}
	\label{fig-network}
\end{figure}

Inspired by the recent success of regression-based object detectors, we propose a new system of neural networks in order to effectively (1) process spatiotemporal information and (2) infer region locations. 
Our methods extends the YOLO deep convolutional neural network into the spatiotemporal domain using recurrent neural networks. 
So, we refer to our method by ROLO (recurrent YOLO).
The architecture of our proposed  ROLO is shown in Fig. \ref{fig-network}. Specifically, 
(1) we use YOLO to collect rich and robust visual features, as well as preliminary location inferences; and we use LSTM in the next stage as it is spatially deep and appropriate for sequence processing. 
(2) Inspired by YOLO’s location inference by regression, we study in this paper the regression capability of LSTM, and propose to concatenate high-level visual features produced by convolutional networks with region information. 
There are three phases for the end-to-end training of the ROLO model: the pre-training phase of convolutional layers for feature learning, the traditional YOLO training phase for object proposal, and the LSTM training phase for object tracking. 

\subsection{Network Training of the Detection Module}

We first pre-train weights with a traditional CNN for general feature learning. The convolutional neural network takes a video frame as its input and produce a feature map of the whole image. 
The convolutional weights are learned with ImageNet data of 1000 classes such that the network has a generalized understanding of almost arbitrary visual objects. During pre-training, the output of the first fully connected layer is a feature vector of size 4096, a dense representation of the mid-level visual features. In theory, the feature vector can be fed into any classification tool (such as an SVM or CNN) to achieve good classification results with proper training.

Once we have the pre-trained weights able to generate visual features, we adopt the YOLO architecture as the detection module. 
On top of the convolutional layers, YOLO adopts fully connected layers to regress feature representation into region predictions. 
These predictions are encoded as an $S \times S \times (B \times 5 + C)$ tensor. It denotes that the image is divided into $S \times S$ splits. Each split has B bounding boxes predicted, represented by its 5 location parameters including $x$, $y$, $w$, $h$, and its confidence $c$. A one-hot feature vector of length $C$ is also predicted, indicating the class label of each bounding box. In our framework, we follow the YOLO architecture and set $S = 7$, $B= 2$, $C= 20$. 
Each bounding box originally consists of 6 predictions: $x$, $y$, $w$, $h$, \textit{class label} and \textit{confidence}, but we nullify class label and confidence for visual tracking, as the evaluation consists of locations only.  
\begin{equation}
B_{t} = (0, x, y, w, h, 0),
\end{equation}
where $(x, y)$ represent the coordinates of the bounding box center relative to the width and the height of the image, respectively. The width and height of the bounding box, are also relative to those of the image. Consequently, $(x, y, w, h) \in [0, 1]$, and it is easier for regression when they are concatenated with the 4096-dimensional visual features, which will be fed into the tracking module.

\subsection{Network Training of the Tracking Module}

At last, we add the LSTM RNNs for the training of the tracking module. There are two streams of data flowing into the LSTMs, namely, the feature representations from the convolutional layers and the detection information $B_{t,i}$ from the fully connected layers. 
Thus, at each time-step $t$, we extract a feature vector of length 4096. We refer to these vectors as $X_{t}$. In addition to $X_{t}$ and $B_{t,i}$, another input to the LSTM is the output of states from the last time-step $S_{t-1}$.
In our objective module we use the Mean Squared Error (MSE) for training:
\begin{equation}
L_{MSE} = \frac{1}{n} \sum_{i=1}^{n} ||B_{target} - B_{pred} ||_{2}^{2},
\end{equation}
where $n$ is the number of training samples in a batch, $y_{pred}$ is the model’s prediction, $y_{target}$ is the target ground truth value and $||\cdot||$ is the squared Euclidean norm. We use the Adam method for stochastic optimization.

\subsection{Alternative Heatmap}
Regressing coordinates directly is highly non-linear and it is difficult for us to interpret the mapping. 
In order to know what really happens in LSTM during tracking, especially under occlusion conditions, we alternatively convert the ROLO prediction location into a feature vector of length 1024, which can be translated into a 32-by-32 heatmap. 
And we concatenate it with the 4096 visual features before feeding into the LSTM. 
The advantage of the heatmap is that it allows to have confidence at multiple spatial locations and we can visualize the intermediate results.  
The heatmap not only acts as an input feature but can also warp predicted positions in the image.
During training, we transfer the region information from the detection box into the heatmap by assigning value $1$ to the corresponding regions while $0$ elsewhere. Specifically, the detection box is converted to be relative to the 32-by-32 heatmap, which is then flattened to concatenate with the 4096 visual features as LSTM input.
Let $H_{target}$ denote the heatmap vector of the groundtruth and $H_{pred}$ denote the heatmap predicted in LSTM output. The objective function is defined as:
\begin{equation}
	L_{MSE} = \frac{1}{n} \sum_{i=1}^{n} ||H_{target} - H_{pred} ||_{2}^{2},
\end{equation} 

\subsection{Spatio-temporal Regression and Spatial Supervision by Region Proposals}
In our findings, LSTM is not only capable of sequence processing but also competent in effective spatio-temporal regression. This regression is two-folds: 
(1) The regression within one unit, i.e., between the visual features and the concatenated region representations. LSTM is capable of inferring region locations from the visual features when they are concatenated to be one unit.
(2) The regression over the units of a sequence, i.e., between concatenated features over a sequence of frames. LSTM is capable of regressing the sequence of features into a predicted feature vector in the next frame. During the regression, LSTM automatically exploits the spatiotemporal information represented by visual features and region locations/heatmaps. 

In the YOLO’s structure, regression in the fully connected layer results in object proposals. They act as soft spatial supervision for the tracking module. The supervision is helpful in two aspects: 
(1) When LSTM interpret the high-level visual features, the preliminary location inference helps to regress the features into the location of a certain visual elements/cues. The spatially supervised regression acts as an online appearance model.
(2) Temporally, the LSTM learns over the sequence units to restrict the location prediction to a spatial range.

\section{Experimental Results} \label{experimental-results}
Our system is implemented in Python using Tensorflow, and runs at 20fps/60fps for YOLO/LSTM respectively, with eight cores of 3.4GHz Intel Core i7-3770 and an NVIDIA TITAN X GPU.   
To aid in reproducing our experiments, we make the source code of our tracker, the pre-trained models, and results available on our \href{http://guanghan.info/projects/ROLO/}{project page}.

\begin{figure}[h] 
	\centering
	\includegraphics[height=1.5in]{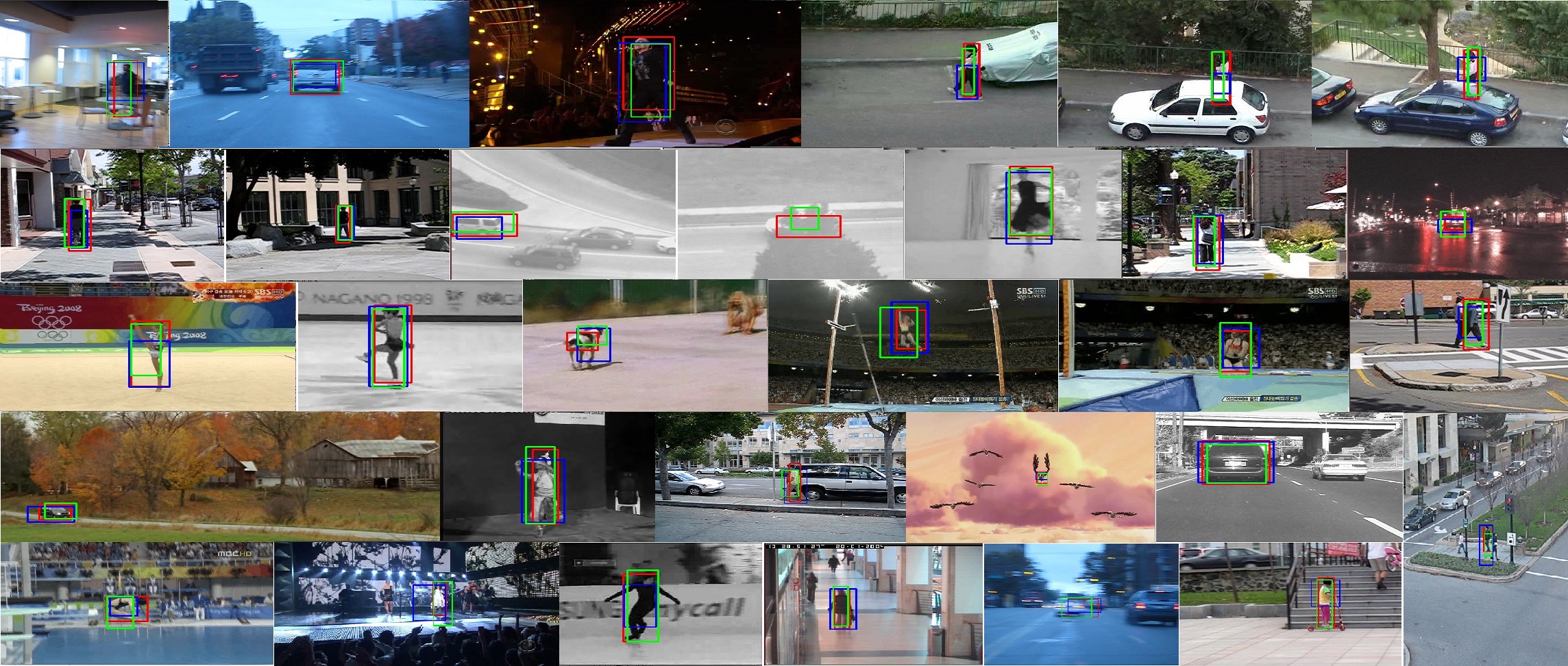}
	\caption{Qualitative tracking results for the 30 test suite of videos. Red boxes are ground truths of the dataset. Green and blue boxes correspond to tracking results for ROLO and YOLO, respectively.}
	\label{fig-qualitative-results}
\end{figure}


Extensive empirical evaluation has been conducted, comparing the performance of ROLO with 10 distinct trackers on a suite of 30 challenging and publicly available video sequences.
Specifically, we compare our results with the top 9 trackers that achieved the best performance evaluated by the benchmark \citep{wu2015object}, including STRUCK \citep{7360205}, CXT \citep{dinh2011context}, OAB \citep{grabner2006real}, CSK \citep{henriques2012exploiting}, VTD \citep{kwon2010visual}, VTS \citep{kwon2011tracking}, LSK \citep{liu2011robust}, TLD \citep{kalal2010pn}, RS \citep{collins2005online}. 
Note that CNN-SVM \citep{hong2015tracking} is another tracking algorithm based on representations from CNN, as a baseline for trackers that adopt deep learning. 
We also use a modified version of SORT \citep{bewley2015sort} to evaluate the tracking performance of YOLO with kalman filter.
As a generic object detector, YOLO can be trained to recognize arbitrary objects. 
Since the performance of ROLO depends on the YOLO part, 
we choose the default YOLO model for fair comparison. The model is pre-trained on ImageNet dataset and finetuned on VOC dataset, 
capable of detecting objects of 20 classes. 
We pick a subset of 30 videos from the benchmark, where the targets belong to these classes.
The video sequences considered in this evaluation are summarized in Table \ref{scores}. According to experimental results of benchmark methods, the average difficulty of OTB-30 is harder than that of the full benchmark.

\subsection{Qualitative Results}
\begin{figure}[h] 
	\tiny
	\captionsetup{justification=centering}
	\includegraphics[width=5.5in]{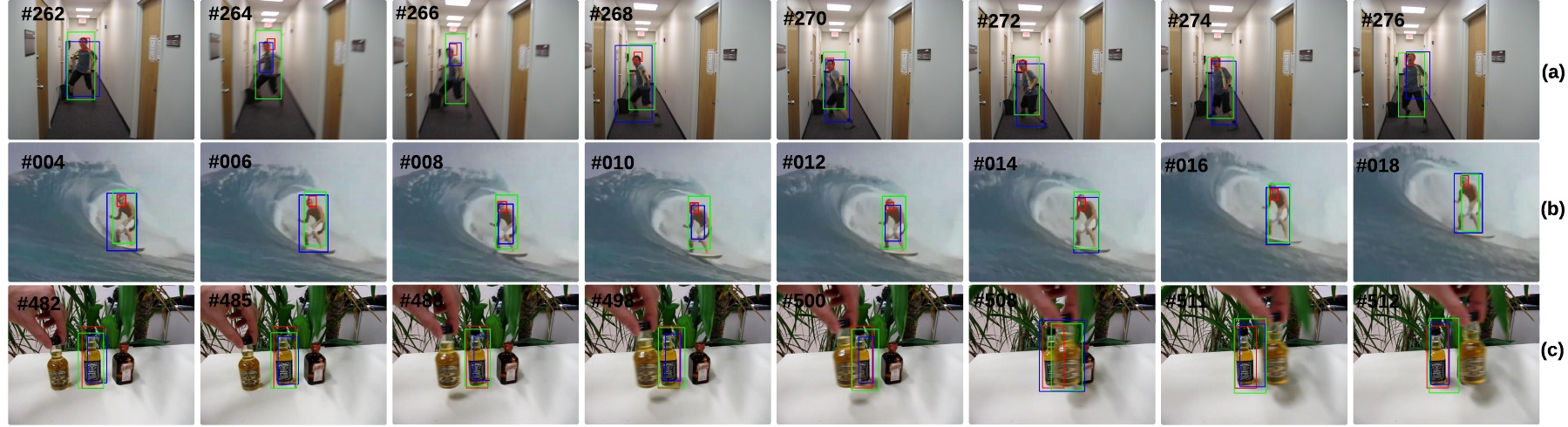}
	\caption{Tracking results for unseen sequences. Red indicates ground truth, while blue and green indicate YOLO detection and ROLO prediction results, respectively.}
	\label{fig-Generalization}
\end{figure}

Since the training data is quite limited, we first test the generalization capability of our neural network.
In Fig. \ref{fig-Generalization}, the model is trained with OTB-30, but tested on unseen video clips. As is shown in Fig. \ref{fig-Generalization} (a)(b), the ground truth of these classes are faces, which does not belong to the pre-trained YOLO classes. In this case, YOLO detects a person as a whole, and ROLO tracks accordingly. Note that when YOLO detection is flawed due to motion blur, ROLO tracking result stays stable with spatio-temporal ponder. In Fig. \ref{fig-Generalization} (c), the object class does belong to the pre-trained YOLO classes but is unseen in any of the training sequences. In this case ROLO tracks it nonetheless.
It proves that: (1) the tracking is generalized to unseen objects, (2) LSTM is capable of interpreting the visual features, and (3) LSTM is able to regress visual features to region inferences with spatial supervision. 
As of interpreting visual features, it is indicated in \citep{dosovitskiy2015inverting} that there is surprisingly rich information contained in these high-level visual features, as the colors and the rough contours of an image can be reconstructed from activations in higher network layers. 
We find the LSTM unit in ROLO interprets visual features and regress them into location predictions, in a way that is similar to the fully connected layers in YOLO. Besides, it renders more stable locations as it considers spatio-temporal history. In contrast to traditional methods for temporal rectification, e.g., the Kalman filter, where the prediction is based solely on the previous locations, ROLO also exploits its history of visual cues. 

The location history in ROLO acts as spatial supervision, which is twofold: (1) when LSTM interpret the high-level visual features, the preliminary location inference helps to regress the features into the location of a certain visual elements/cues. The spatially supervised regression acts as an online appearance model (2) temporally, the LSTM learns over the sequence units to restrict the location prediction to a spatial range.   

\subsection{Handling Occlusions}
\begin{figure}[h] 
	\centering
	\captionsetup{justification=centering}
	\includegraphics[width=5.5in]{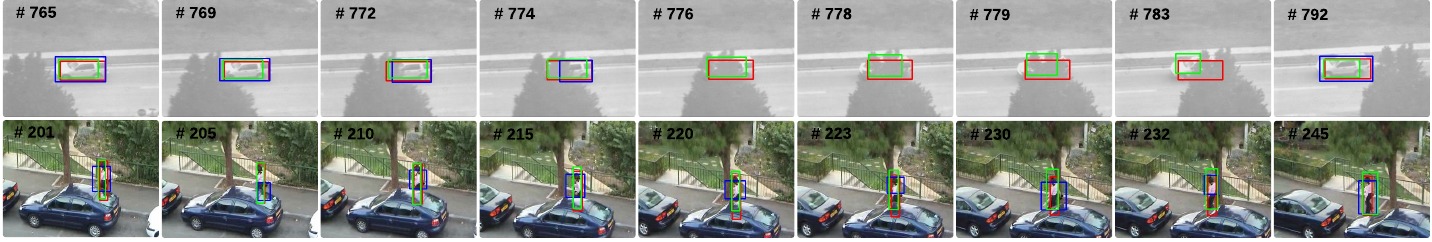}
	\caption{Spatio-temporal robustness against occlusion in unseen frames.}
	\label{fig-occlusion}
\end{figure}

Qualitative result in Fig.\ref{fig-occlusion} shows that ROLO successfully tracks the object under occlusion challenges in unseen frames. Note that during frames 776-783, ROLO continues tracking the vehicle even though the detection module fails. 
\begin{figure}[h] 
	\centering
	\captionsetup{justification=centering}
	\includegraphics[width=5in]{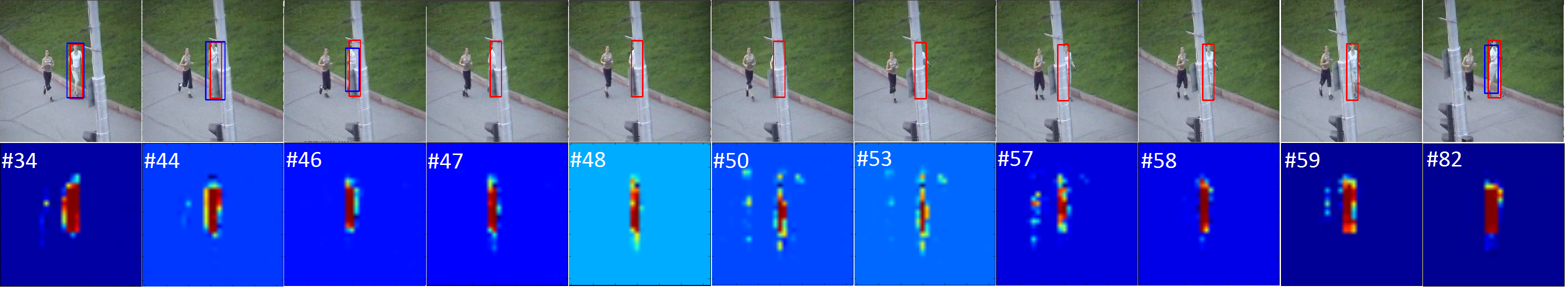}
	\caption{Robustness against occlusion in unseen video clip. Results are shown in heatmap. Blue and Red bounding boxes indicate YOLO detection and the ground truth, respectively.}
	\label{fig-occlusion-heatmap}
\end{figure}
We also train an alternative ROLO model with heatmap instead of location coordinates, in order to analyze LSTM under occlusion conditions. The model is trained offline with 1/3 frames from OTB-30 tested on unseen videos. 
It is shown in Fig. \ref{fig-occlusion-heatmap} that ROLO tracks the object in near-complete occlusions.  
Even though two similar targets simultaneously occur in this video, ROLO tracks the correct target as the detection module inherently feeds the LSTM unit with spatial constraint.
Note that between frame 47-60, YOLO fails in detection but ROLO does not lose the track. 
The heatmap is involved with minor noise when no detection is presented as the similar target is still in sight. Nevertheless, ROLO has more confidence on the real target even when it is fully occluded, 
as ROLO exploits its history of locations as well as its visual features.  
ROLO is proven to be effective due to several reasons: (1) the representation power of the high-level visual features from the convNets, (2) the feature interpretation power of LSTM, therefore the ability to detect visual objects, which is spatially supervised by a location or heatmap vector, (3) the capability of regressing effectively with spatio-temporal information.

\subsection{Quantitative Results}

\begin{figure}[!h] 
	\centering
	\captionsetup{justification=centering}
	\includegraphics[width=5.2in]{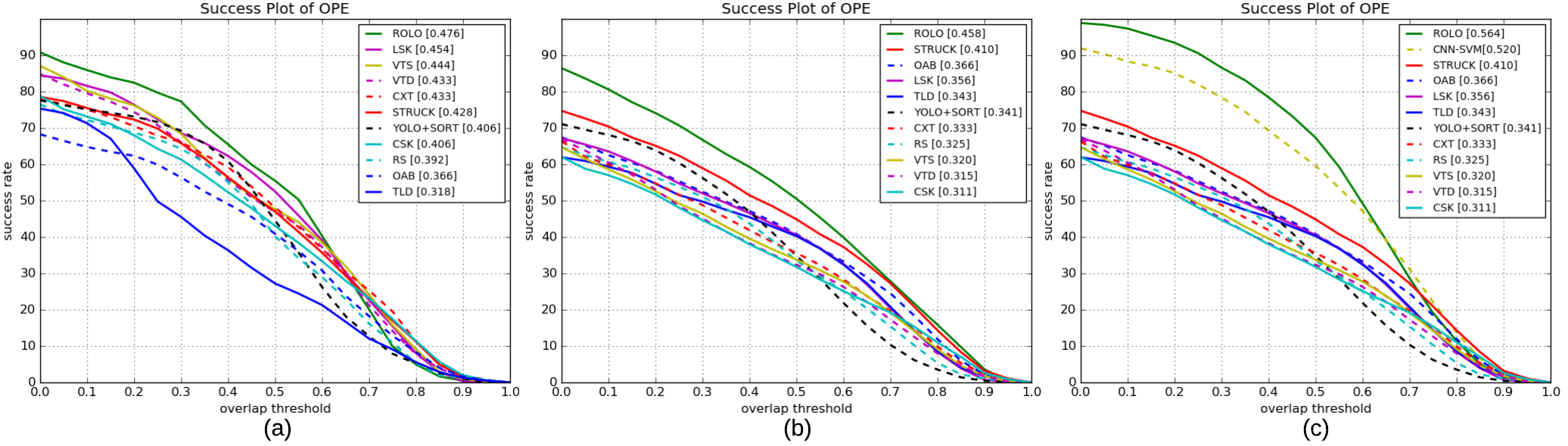}
	\caption{Success Plots of OPE (one pass evaluation) on the 3 conducted experiments.}
	\label{fig-OPEs}
\end{figure}

\begin{table}[t]
	\centering
	\tiny
	\caption{Summary of Average Overlap Scores (AOS) results for all 12 trackers. The best and second best results are in {\color[HTML]{32CB00} \textbf{green}} and {\color[HTML]{FE0000} \textbf{red}} colors, respectively.}
	\label{scores}
	\begin{tabular}{|l|l|l|l|l|l|l|l|l|l|l|l|l|}
		\hline
		Sequence & ROLO & \begin{tabular}[c]{@{}l@{}}YOLO\\ +SORT\end{tabular} & \begin{tabular}[c]{@{}l@{}}STR\\ UCK\end{tabular} & CXT & TLD & OAB & CSK & RS & LSK & VTD & VTS & \begin{tabular}[c]{@{}l@{}}CNN-\\ SVM\end{tabular} \\ \hline
		Human2 & {\color[HTML]{333333} 0.545} & {\color[HTML]{FE0000} \textbf{0.636}} & {\color[HTML]{32CB00} \textbf{0.652}} & 0.248 & 0.398 & 0.611 & 0.169 & 0.524 & 0.438 & 0.185 & 0.185 & 0.617 \\ \hline
		Human9 & {\color[HTML]{32CB00} \textbf{0.352}} & {\color[HTML]{333333} 0.193} & 0.065 & 0.062 & 0.159 & 0.170 & 0.220 & 0.060 & 0.291 & 0.244 & 0.111 & {\color[HTML]{FE0000} \textbf{0.350}} \\ \hline
		Gym & {\color[HTML]{32CB00} \textbf{0.599}} & {\color[HTML]{FE0000} \textbf{0.460}} & 0.350 & 0.423 & 0.276 & 0.069 & 0.219 & 0.413 & 0.101 & 0.367 & 0.359 & 0.438 \\ \hline
		Human8 & {\color[HTML]{333333} 0.364} & {\color[HTML]{333333} 0.416} & 0.127 & 0.109 & 0.127 & 0.095 & 0.171 & 0.333 & {\color[HTML]{32CB00} \textbf{0.653}} & 0.246 & 0.336 & {\color[HTML]{FE0000} \textbf{0.452}} \\ \hline
		Skater & {\color[HTML]{32CB00} \textbf{0.618}} & 0.283 & 0.551 & {\color[HTML]{FE0000} \textbf{0.584}} & 0.326 & 0.481 & 0.431 & 0.575 & 0.481 & 0.471 & 0.470 & 0.571 \\ \hline
		SUV & {\color[HTML]{333333} 0.627} & 0.455 & 0.519 & {\color[HTML]{FE0000} \textbf{0.715}} & 0.660 & 0.619 & 0.517 & 0.341 & 0.583 & 0.431 & 0.468 & {\color[HTML]{32CB00} \textbf{0.724}} \\ \hline
		BlurBody & 0.519 & 0.337 & {\color[HTML]{32CB00} \textbf{0.696}} & {\color[HTML]{FE0000} \textbf{0.663}} & 0.391 & 0.671 & 0.381 & 0.277 & 0.264 & 0.238 & 0.238 & 0.603 \\ \hline
		CarScale & {\color[HTML]{FE0000} \textbf{0.565}} & {\color[HTML]{32CB00} \textbf{0.627}} & 0.350 & 0.620 & 0.434 & 0.325 & 0.400 & 0.272 & 0.510 & 0.442 & 0.436 & 0.394 \\ \hline
		Dancer2 & 0.627 & 0.201 & {\color[HTML]{32CB00} \textbf{0.776}} & 0.707 & 0.651 & 0.766 & {\color[HTML]{FE0000} \textbf{0.776}} & 0.721 & 0.751 & 0.704 & 0.717 & 0.758 \\ \hline
		BlurCar1 & 0.537 & 0.082 & {\color[HTML]{FE0000} \textbf{0.760}} & 0.187 & 0.605 & {\color[HTML]{32CB00} \textbf{0.780}} & 0.011 & 0.566 & 0.475 & 0.210 & 0.210 & 0.743 \\ \hline
		Dog & {\color[HTML]{333333} 0.429} & 0.241 & 0.264 & {\color[HTML]{32CB00} \textbf{0.592}} & {\color[HTML]{FE0000} \textbf{0.569}} & 0.317 & 0.308 & 0.326 & 0.080 & 0.302 & 0.299 & 0.315 \\ \hline
		Jump & {\color[HTML]{32CB00} \textbf{0.547}} & {\color[HTML]{FE0000} \textbf{0.208}} & 0.105 & 0.056 & 0.070 & 0.085 & 0.094 & 0.050 & 0.132 & 0.057 & 0.053 & 0.069 \\ \hline
		Singer2 & {\color[HTML]{FE0000} \textbf{0.588}} & {\color[HTML]{333333} 0.400} & 0.040 & 0.052 & 0.026 & 0.045 & 0.043 & 0.067 & 0.073 & 0.416 & 0.332 & {\color[HTML]{32CB00} \textbf{0.675}} \\ \hline
		Woman & {\color[HTML]{333333} 0.649} & 0.358 & {\color[HTML]{FE0000} \textbf{0.693}} & 0.208 & 0.129 & 0.466 & 0.194 & 0.354 & 0.140 & 0.145 & 0.132 & {\color[HTML]{32CB00} \textbf{0.731}} \\ \hline
		David3 & {\color[HTML]{333333} 0.622} & 0.224 & 0.279 & 0.117 & 0.096 & 0.318 & {\color[HTML]{333333} 0.499} & {\color[HTML]{32CB00} \textbf{0.731}} & 0.381 & 0.372 & 0.541 & {\color[HTML]{FE0000} \textbf{0.714}} \\ \hline
		Dancer & {\color[HTML]{32CB00} \textbf{0.755}} & 0.551 & 0.625 & 0.623 & 0.394 & 0.604 & 0.609 & 0.489 & 0.589 & {\color[HTML]{333333} 0.720} & {\color[HTML]{FE0000} \textbf{0.728}} & 0.645 \\ \hline
		Human7 & {\color[HTML]{333333} 0.456} & {\color[HTML]{333333} 0.291} & {\color[HTML]{333333} 0.466} & 0.429 & {\color[HTML]{32CB00} \textbf{0.675}} & 0.421 & 0.350 & 0.252 & 0.371 & 0.299 & 0.206 & {\color[HTML]{FE0000} \textbf{0.482}} \\ \hline
		Bird1 & {\color[HTML]{32CB00} \textbf{0.362}} & 0.048 & 0.093 & 0.014 & 0.004 & 0.055 & 0.018 & {\color[HTML]{333333} 0.225} & 0.032 & 0.023 & 0.023 & {\color[HTML]{FE0000} \textbf{0.240}} \\ \hline
		Car4 & {\color[HTML]{32CB00} \textbf{0.768}} & {\color[HTML]{FE0000} \textbf{0.690}} & 0.436 & 0.234 & 0.537 & 0.180 & 0.352 & 0.082 & 0.141 & 0.400 & 0.392 & 0.480 \\ \hline
		CarDark & 0.674 & 0.211 & {\color[HTML]{32CB00} \textbf{0.872}} & 0.540 & 0.423 & 0.765 & 0.744 & 0.334 & {\color[HTML]{FE0000} \textbf{0.800}} & 0.521 & 0.717 & 0.747 \\ \hline
		Couple & {\color[HTML]{333333} 0.464} & 0.204 & 0.484 & 0.464 & {\color[HTML]{009901} \textbf{0.761}} & 0.346 & 0.074 & 0.054 & 0.073 & 0.068 & 0.057 & {\color[HTML]{FE0000} \textbf{0.612}} \\ \hline
		Diving & {\color[HTML]{32CB00} \textbf{0.659}} & {\color[HTML]{333333} 0.166} & 0.235 & 0.196 & 0.180 & 0.214 & 0.235 & {\color[HTML]{FE0000} \textbf{0.346}} & 0.214 & 0.210 & 0.213 & 0.259 \\ \hline
		Human3 & {\color[HTML]{32CB00} \textbf{0.568}} & {\color[HTML]{333333} 0.386} & 0.007 & 0.009 & 0.007 & 0.010 & 0.011 & 0.038 & 0.022 & 0.018 & 0.018 & {\color[HTML]{FE0000} \textbf{0.540}} \\ \hline
		Skating1 & {\color[HTML]{32CB00} \textbf{0.572}} & 0.443 & 0.285 & 0.103 & 0.184 & 0.368 & 0.478 & 0.269 & 0.472 & 0.492 & {\color[HTML]{FE0000} \textbf{0.482}} & 0.402 \\ \hline
		Human6 & {\color[HTML]{32CB00} \textbf{0.532}} & {\color[HTML]{FE0000} \textbf{0.376}} & 0.217 & 0.159 & 0.282 & 0.207 & 0.208 & 0.183 & 0.363 & 0.168 & 0.168 & 0.200 \\ \hline
		Singer1 & {\color[HTML]{FE0000} \textbf{0.653}} & 0.332 & 0.312 & 0.413 & {\color[HTML]{32CB00} \textbf{0.684}} & 0.345 & 0.313 & 0.337 & 0.284 & {\color[HTML]{333333} 0.464} & 0.460 & 0.340 \\ \hline
		Skater2 & {\color[HTML]{32CB00} \textbf{0.652}} & {\color[HTML]{333333} 0.532} & 0.536 & 0.341 & 0.263 & 0.500 & 0.546 & 0.280 & 0.416 & 0.454 & 0.454 & {\color[HTML]{FE0000} \textbf{0.564}} \\ \hline
		Walking2 & {\color[HTML]{32CB00} \textbf{0.595}} & 0.362 & {\color[HTML]{FE0000} \textbf{0.500}} & 0.394 & 0.299 & 0.359 & 0.492 & 0.290 & 0.421 & 0.354 & 0.360 & 0.479 \\ \hline
		BlurCar3 & 0.539 & 0.191 & {\color[HTML]{FE0000} \textbf{0.780}} & 0.574 & 0.639 & {\color[HTML]{333333} 0.720} & 0.430 & 0.276 & 0.644 & 0.188 & 0.188 & {\color[HTML]{32CB00} \textbf{0.793}} \\ \hline
		Girl2 & {\color[HTML]{333333} 0.517} & 0.337 & 0.227 & 0.169 & 0.070 & 0.071 & 0.060 & {\color[HTML]{32CB00} \textbf{0.687}} & 0.494 & 0.257 & 0.257 & {\color[HTML]{FE0000} \textbf{0.681}} \\ \hline
	\end{tabular}
\end{table}

\begin{figure}[!ht]
	\centering
	\subfloat[\label{subfig-1:robust}]{%
		\includegraphics[width=0.35\textwidth]{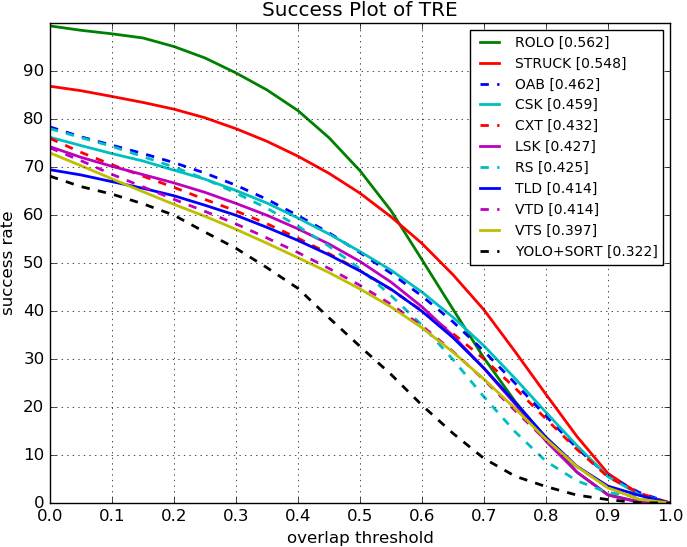}
	}
	\subfloat[\label{subfig-2:robust}]{%
		\includegraphics[width=0.35\textwidth]{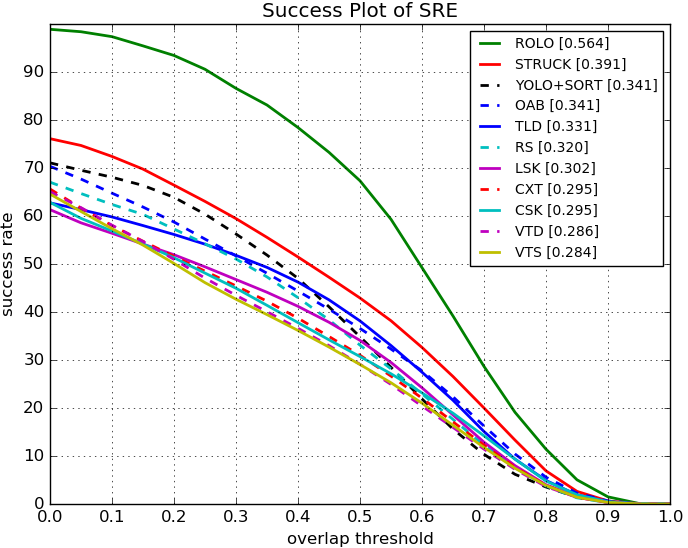}
	}
	\caption{Success Plots for TRE (temporal robustness evaluation) and SRE (spatial robustness evaluation) on the OTB-30 benchmark.}
	\label{fig-robustness-eval}
\end{figure}
We first train an LSTM model with 22 videos in OTB-30 and test the rest 8 clips. 
The OPE result is shown in Fig \ref{fig-OPEs}(a). It again demonstrates the generalization ability of LSTM regression, but it also shows that 
the model does not perform extremely well
when limited dynamics are used for training.
In order to learn whether training with similar dynamics can improve performance, we train a 2nd LSTM model with 1/3 frames and their ground-truth labels of OTB-30, testing on the whole sequence frames. The OPE result is shown in \ref{fig-OPEs}(b).   
We find that, once trained on auxiliary frames with the similar dynamics, ROLO will perform better on testing sequences. This attribute makes ROLO especially useful in surveillance environments, where models can be trained offline with pre-captured data. 
Considering this attribute, we experiment incrementing training frames, expecting to see an improved performance. We train a 3rd LSTM model with 1/3 ground truths, but with all the sequence frames. Results in Fig \ref{fig-OPEs}(c) show that even without addition of ground truth boxes, the performance can increase dramatically when more frames are used for training to learn the dynamics. 
It also shows that for tracking, the training data in the benchmark is quite limited \citep{nam2015learning}. 
Its SRE and TRE results are shown in Fig. \ref{fig-robustness-eval} for robustness evaluation.
The AOS for each video sequence is illustrated in Table \ref{scores}. Our method achieves the best performance for most test video sequences, often outperforms the second best by a large margin.

\subsection{Parameter Sensitivity and Tracker Analysis}
CNN-SVM is CNN-based tracking method with robust features, but lacks temporal information to deal with severe occlusion. 
YOLO with kalman filter takes into account the temporal evolution of locations, while ignorant of actual environments. Due to fast motions, occlusions, and therefore occasionally poor detections, YOLO with the kalman filter perform inferiorly lacking knowledge of the visual context.
In contrast, with LSTM ROLO synthesizes over sequences the robust image features as well as their soft spatial supervision.  
ROLO is spatially deep, as it is capable of interpreting the visual features and detecting objects on its own, which can be spatially supervised by concatenating locations or heatmaps to the visual features. It is also temporally deep by exploring temporal features as well as their possible locations. 
\begin{figure}[!ht]
	\centering
	\subfloat[\label{subfig-1:steps}]{%
		\includegraphics[width=0.35\textwidth]{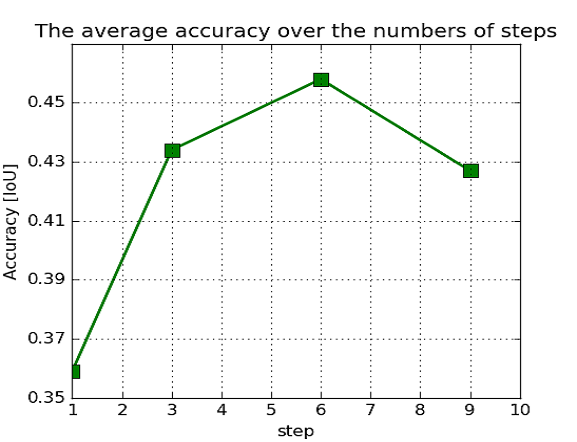}
	}
	\subfloat[\label{subfig-2:steps}]{%
		\includegraphics[width=0.35\textwidth]{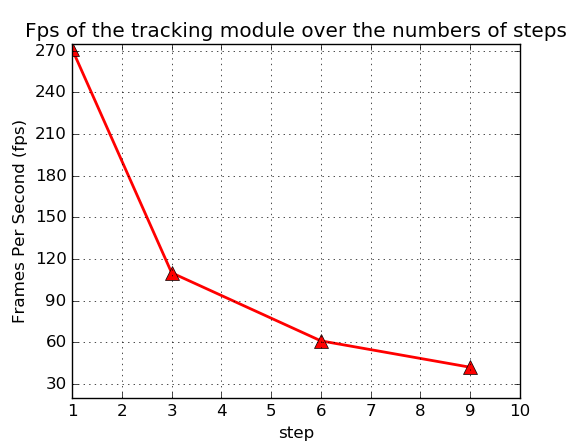}
	}
	\caption{Average IOU scores and fps under various step sizes.}
	\label{fig-steps}
\end{figure}
Step size
denotes the number of previous frames considered each time for a prediction by LSTM. In previous experiments, we used 6 as the step number. 
In order to shed light upon how sequence step of LSTM affects the overall performance and running time, we repeat the 2nd experiment with various step sizes, and illustrate the results in Fig. \ref{fig-steps}. 
In our experiments, we also tried dropouts on visual features, random offset of detection boxes during training intended for more robust tracking, and auxiliary cost to the objective function to emphasize detection over visual features, but these results are inferior to what is shown.

\section{Conclusion and Future Work}

In this paper, we have successfully developed a new method of spatially supervised recurrent convolutional neural networks for visual object tracking.  Our proposed ROLO method extends the deep neural network learning and analysis into the spatiotemporal domain. 
We have also studied LSTM's interpretation and regression capabilities of high-level visual features. Our proposed tracker is both spatially and temporally deep, and can effectively tackle problems of major occlusion and severe motion blur.
Our extensive experimental results and performance comparison with state-of-the-art tracking methods on challenging benchmark tracking datasets shows that our tracker is more accurate and robust while maintaining low computational cost. For most test video sequences, our method achieves the best tracking performance, often outperforms the second best by a large margin. 

In  our future research, we will study two stacked LSTMs for the optimization of cost functions on heatmaps and locations individually, which may provide more room for further performance improvement. 
We will focus on efficient online learning, in order to maintain high performance while tracking an object in unseen dynamics with real-time performance. We will also explore data association techniques in order for ROLO to work for multi-target tracking purposes.

%

\small
\bibliography{ROLO}

\end{document}